# S-R2F2U-Net: A single-stage model for teeth segmentation

Mrinal Kanti Dhar[1,*] and Mou Deb[2]

[1]Department of Computer Science, University of Wisconsin-Milwaukee, WI, USA
[2]Department of Electrical Engineering, South Dakota School of Mines and Technology, SD, USA

[1,*]mdhar@uwm.edu and [2]mou.deb@mines.sdsmt.edu

## ABSTRACT

Precision teeth segmentation is crucial in the oral sector because it provides location information for orthodontic therapy, clinical diagnosis, and surgical treatments. In this paper, we investigate residual, recurrent, and attention networks to segment teeth from panoramic dental images. Based on our findings, we suggest three single-stage models: Single Recurrent R2U-Net (S-R2U-Net), Single Recurrent Filter Double R2U-Net (S-R2F2U-Net), and Single Recurrent Attention Enabled Filter Double (S-R2F2-Attn-U-Net). Particularly, S-R2F2U-Net outperforms state-of-the-art models in terms of accuracy and dice score. A hybrid loss function combining the cross-entropy loss and dice loss is used to train the model. In addition, it reduces around 45% of model parameters compared to the R2U-Net model. Models are trained and evaluated on a benchmark dataset containing 1500 dental panoramic X-ray images. S-R2F2U-Net achieves 97.31% of accuracy and 93.26% of dice score, showing superiority over the state-of-the-art methods. Codes are available at https://github.com/mrinal054/teethSeg_sr2f2u-net.git.

## Introduction

In the oral field, precise teeth segmentation is critical because it offers location information for orthodontic treatment, clinical diagnosis, and surgical procedures. To identify humans using dental radiographic images, teeth should be segmented in collecting tooth shape information and attributes. Other applications of tooth segmentation include dental structure detection, missing teeth detection, teeth numbering, implant planning, age, and forensic identification. But segmenting teeth manually is in fact a time-consuming and laborious task. Even semi-automatic methods also require dentists' prior knowledge and intervention. It becomes more challenging even for professionals to correctly identify some tooth regions when it is a relatively low-resolution image.

Panoramic X-ray images are useful in the diagnosis of dental diseases during clinical examination as it allows visualization of dental and buccal irregularities. However, such images lack the precise anatomic detail that intraoral periapical radiographs provide. Sometimes it suffers from low contrast, noise, magnification, geometric distortion, or overlapped images of teeth. The presence of bones like the jaw and noise in the radiograph make the segmentation task more challenging. Besides, there has been a scarcity of benchmark datasets with dental radiography until Silva et al.[1] created a dataset that consists of 1500 dental panoramic X-ray images.

Teeth segmentation methods can be divided into two categories – traditional methods which incorporate prior knowledge and image features and deep learning-based methods. Some of the popular techniques in traditional methods are shape model-based, threshold and region-based, templating matching-based, and boundary-based. Lira et al.[2] proposed a segmentation model that consists of quadtree decomposition, morphological operations, thresholding, and shape analysis (PCA). No quantitative result is reported in their paper. Grafova et al.[3] proposed a statistical measures-based method to detect edges in dental panoramic X-ray images that includes selecting the type of edge detector from *N* edge detection results and computing correspondence maps that are constructed from those *N* results. However, their approach is time-consuming as the computational time for determining parameters is quite high. Indraswari et al.[4] proposed a segmentation technique that generates directional images using Decimation-Free Directional Filter Bank Thresholding (DDFBT), and then segments using Multistage Adaptive Thresholding (MAT) with Sauvola Local Thresholding. Poonsri et al.[5] proposed a template matching method with the Otsus threshold and Mahalanobis distance technique to segment teeth. However, the process is unreliable as finding the tooth roots using the thresholding strategy is quite difficult.

Deep learning has grown in popularity with AlexNet's[6] success in the 2012 Imagenet large-scale visual recognition challenge. According to a study, the number of deep learning-based research papers for medical image segmentation increased rapidly from 2014 to 2018[7]. Several deep learning approaches have been proposed since 2012. AlexNet, VGG[8], GoogleNet[9], ResNet[10], DenseNet[11], fully convolutional neural networks (FCNN)[12], U-Net[13], and others are some of the most popular models. As deep learning methods are mostly data-driven, they require a good amount of data to train the model. Most of the datasets used in teeth segmentation from dental panoramic images were small-sized until Silva et al.[1] created a dataset that consists of 1500 images. Jader et al.[14] and Lee et al.[15] used masked R-CNN to segment teeth from dental panoramic X-ray images. Wirtz et al.[16] proposed a 2D coupled shape model that uses U-Net to generate binary masks of the teeth area. Koch et al.[17] applied fully convolutional neural networks (FCN) on U-Net architecture. They used six configurations based on different loss functions and achieved the highest dice score for 3 class cross-entropy along with bootstrapping and horizontal reflections. Zhao et al.[18] proposed an attention-based two-stage network called TSASNet. The first stage is a pixel-wise contextual attention network, whereas the second stage performs segmentation on the attention map achieved from the attention network. Both Zhao et al. and Koch et al. used the dataset from Silva et al.[1] to train and evaluate their model. Pinheiro et al.[19] incorporate the PointRend module with mask R-CNN to improve instance segmentation. Chen et al.[20] extended the multi-scale spatial pyramid pooling (SPP) to implement their multi-scale location perception network (MSLPNet). In addition, they proposed a novel multi-scale structural similarity (MS-SSIM) loss function. Salih and Kevin[21] proposed a method called local ternary encoder-decoder neural network (LTPEDN) where they replaced the LBC layers in local binary convolutional-deconvolutional neural network (LBCDNN) with LTP layers.

In this paper, our contribution can be summarized as follows –

1. We segment teeth from panoramic dental X-ray images using 4 state-of-the-art methods (Attention U-Net[22], U$^2$ Net[23], R2U-Net[24], and ResUNet-a[25]) to examine the effects of residual, recurrent, and attention networks.
2. Based on our findings we propose 3 novel single-stage methods – Single Recurrent R2U-Net (S-R2U-Net), Single Recurrent Filter Double R2U-Net (S-R2F2U-Net), and Single Recurrent Attention Enabled Filter Double (S-R2F2-Attn-U-Net). These architectures not only achieve prominent dice scores but also reduce the number of model parameters compared to the original R2U-Net architecture.
3. We use a hybrid loss function comprising cross-entropy loss and dice loss that boosts the model performance.
4. Instead of implementing separate architectures, we develop a switch-based single U-Net architecture that can accommodate as many recurrent, residual, attention, and filter-doubling modules as the user requires. Though we investigate three configurations in this paper, many other configurations can also be generated from our implementation.
5. Our proposed S-R2F2U-Net model surpasses state-of-the-art approaches with a 97.31% accuracy and a 93.26% dice score.

## Materials and Method

**Data.** We use the benchmark public dataset to evaluate our frameworks. It consists of 1500 dental panoramic X-ray images divided into 10 categories. The summary of the dataset is shown in Supplementary Table S1. With around a 60:10:30 split, we divided the dataset into training, validation, and testing. All images are taken randomly. In total, there are 911, 153, and 436 images for training, validation, and testing, respectively. The original image size is 1991 by 1127 pixels. So, each image is split into small patches of size $512 \times 512$. A $10 \times 10$ overlap is considered between adjacent patches while sliding the patch window. All images are normalized to the range [0, 1].

**Switches.** We implement a switch-based U-Net architecture as shown in Figure 1(A). Switches are used to activate recurrent, residual, attention, and filter doubling modules individually. There are 5 switches – $SW_C$, $SW_R$, $SW_{RS}$, $SW_A$, and $SW_{FD}$ that trigger Convolution-BN-ReLU (CBR), recurrent, residual, attention, and filter doubling blocks, respectively. Both $SW_C$ and $SW_R$ can't be triggered together, instead, one of them will activate at a time, thus enabling activation of either the CBR block or attention block.

**Configuring recurrent convolutional layer (RCL).** A convolutional layer receives an image or some input feature maps and produces corresponding output feature maps. Mathematically, if it receives an input $u$ of $k$ feature maps from the $(l-1)^{th}$, the output at each channel of the $l^{th}$-layer is computed as follows:

$$O^{[l]} = \sigma^{[l]}\left(z^{[l]}\left(w^{[l]}, u^{[l-1]}\right)\right) \tag{1}$$

Where $w$ and $b$ are learned weights and biases. $\sigma(\cdot)$ is a nonlinear activation function, and $z(\cdot)$ is a linear operator described as –

$$z^{[l]}(w, u) = \sum_k \left(w^{[l]}\right)^T u_k^{[l-1]} + b_k^{[l]} \tag{2}$$

In Recurrent-CNN, convolutional architecture in an RCL unit is developed over discrete time steps. Each unit's activity is regulated by the activities of its neighboring units, although the input is static. Such evolvement improves the model's capacity to assimilate context information. So, now in addition to the standard CNN unit, a recurrent connection is induced in it. If $(i,j)$ be the pixel location of the $k^{th}$ feature map of the $l^{th}$ layer, then the linear output of the RCL unit at time step $t$ can be expressed as:

$$z_{ijk}^{[t]}(w, u, x) = \left(w_k^f\right)^T u_{i,j}^{[t]} + (w_k^r)^T x_{i,j}^{[t-1]} + b_k \tag{3}$$

Where $u$ and $x$ represent feedforward and recurrent input, respectively, $w_k^f$ and $w_k^r$ represent vectorized feed-forward weights and recurrent weights, respectively. The fact that, unlike equation (1) in which the CNN unit is stated in terms of layers, now the equation (3) evolves over time. A non-linear activation function is then applied to this linear output. In this paper, we have used rectified linear unit (ReLU) as the non-linear activation function which can be expressed as follows:

$$\mathcal{R}(\boldsymbol{w}, \boldsymbol{u}, \boldsymbol{x}) = (\sigma\left(z_{ijk}^{[t]}\right) = max\left(z_{ijk}^{[t]}, 0\right) \tag{4}$$

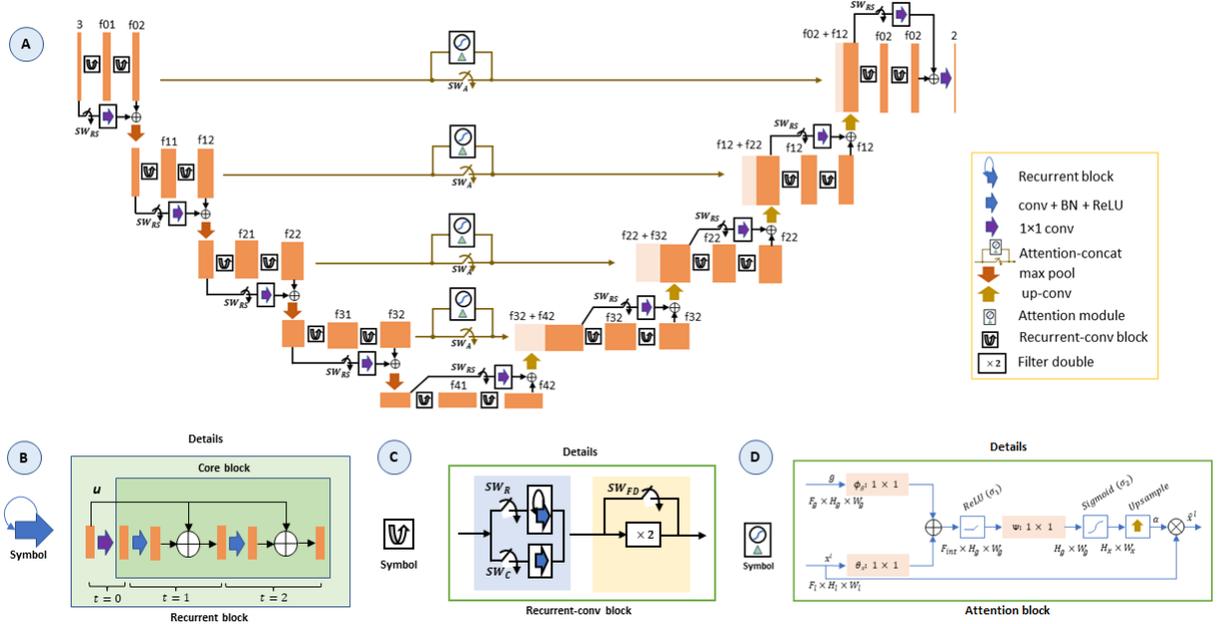

**Figure 1.** Segmentation model. (A) Switch-based U-Net architecture. (B) Recurrent module shown for *t=2*. (C) Switching between recurrent and Conv-BN-ReLU (CBR) blocks. Filter doubling is also embedded. It doubles the input filters. (D) Attention module. More details of modules B-D are given in Supplementary Figure S1.

The detail of the recurrent module is shown in Figure 1(B) where the RCL unit is unfolded over time $t = 2$. We see that though recurrent inputs $\boldsymbol{x}$ evolve over iterations the feedforward unit $\boldsymbol{u}$ remains unchanged. Outputs are generated by adding the recurrent connection from the previous time step and the feedforward connection. As shown in Figure 1(C), feature maps can be generated either by Conv-BN-ReLU (CBR) block that avoids recurrent using switch $SW_C$ or by recurrent block using switch $SW_R$.

**Configuring residual connection.** Going through non-linear activation functions while training a deeper neural network may rise to exploding gradient and vanishing gradient problems. A residual connection achieves better convergence by skipping some of the layers and connecting the input to some latter parts of the network through identity mapping or linear transformation. In this paper, each level of the decoder or encoder path has 3 sets of feature maps. In the proceeding sections, $s_{LA}$ is used to describe the feature maps, where $L$ is the level index and $A$ is the set index. The residual output can be described as follows –

$$C(s_{L0}) + s_{L2} = C(s_{L0}) + \mathcal{R}(\boldsymbol{w}, s_{L1}, \boldsymbol{x}) \tag{5}$$

Where $C(\cdot)$ denotes $1 \times 1$ convolution. $s_{L0}$ is the input feature map of the $L^{th}$ level, whereas $s_{L1}$ and $s_{L2}$ are two subsequent output feature maps after performing some convolution or recurrent operations. In Figure 1, switch $SW_{RS}$ activates residual connections.

**Configuring attention gate.** Oktay et. al.[22] used attention gate modules ($AGs$) in U-Net architecture for medical image segmentation. As shown in Figure 1(D), the attention gate calculates the attention coefficient ($\alpha$) which is the result of the additive attention computed from the input feature maps ($x$) and contextual information provided by the gating signal ($g$). By paying attention to task-relevant activations, the attention coefficient identifies important spatial regions. The output $\hat{x}_{i,c}^{[l]}$ of the $AG$ for each pixel $i$ in $c$ channel at $l^{th}$-layer is computed as follows –

$$I_{map}^{[l]} = \theta_x^T x_i^{[l]} + \phi_g^T g_i + b_g \tag{6}$$

$$I_{att}^{[l]} = \sigma_1\left(I_{map}^{[l]}\right) \tag{7}$$

$$q_{att}^{[l]} = \psi^T\left(I_{att}^{[l]}\right) + b_\psi \tag{8}$$

| Attribute Configuration | Residual | Recurrent1 | Recurrent2 | Filter doubling | Attention |
|---|---|---|---|---|---|
| Attention U-Net | ✗ | ✗ | ✗ | ✗ | ✓ |
| R2U-Net | ✓ | ✓ | ✓ | ✗ | ✗ |
| S-R2U-Net | ✓ | ✗ | ✓ | ✗ | ✗ |
| S-R2F2U-Net | ✓ | ✗ | ✓ | ✓ | ✓ |
| S-R2F2-Attn-U-Net | ✓ | ✗ | ✓ | ✓ | ✓ |

**Table 1.** Configurations of different models. Recurrent1 and Recurrent2 are the recurrent modules in each level.

$$\alpha_i^{[l]} = \sigma_2\left(q_{att}^{[l]}\right) \quad (9)$$

$$\hat{x}_{i,c}^{[l]} = x_{i,c}^{[l]} * \alpha_i^{[l]} \quad (10)$$

Where $I_{map}^l$ is an intermediate mapping function that maps input feature maps and the gating signal to some intermediate feature maps. In this paper, the number of filters is chosen as half the number of input filters. $I_{att}^l$ and $q_{att}^l$ are the intermediate attention and final addition attention, respectively. $\sigma_1$ and $\sigma_2$ are two non-linear activation functions. We use ReLU and sigmoid activation functions as $\sigma_1$ and $\sigma_1$, respectively. $\theta_x \in \mathbb{R}^{f_l \times f_{int}}$, $\phi_g \in \mathbb{R}^{f_g \times f_{int}}$, and $\psi \in \mathbb{R}^{f_{int} \times 1}$ are learned weights. $b_g$ and $b_\psi$ are biases. The operator $*$ indicates element-wise multiplication.

**Filter doubling.** In the encoder path, switch $SW_{FD}$ enables a filter doubling option. Consider, $f_{L0}$, $f_{L1}$, and $f_{L2}$ be the no. of feature maps in $s_{L0}$, $s_{L1}$, and $s_{L2}$, respectively. If filter doubling is enabled, then $f_{L2} = 2 \times f_{L1}$. We find that filter doubling not only achieves better performance but also reduces the network size as smaller base filters are required then. Particularly, if $s_{L1}$ is achieved without recurrent block, then filter doubling is very effective in achieving a decent dice score. But to introduce filter doubling in residual-recurrent architecture, we need some modifications as the no. of input and output feature maps is different now.

(i) For the recurrent module, $s_{L1}$ is the input to the subsequent recurrent module, and $f_{L1}$ and $f_{L2}$ are different now. So, while applying $C(\cdot)$ to $s_{L1}$, no. of filters must be adjusted accordingly.
(ii) A similar adjustment is required for the additive residual connection. Though in this paper, we use 3 sets of feature maps in each encoder or decoder level, in practice a user can have different sets of feature maps at the same level. So, to generalize the network, we define no. of filters while applying $C(\cdot)$ to $s_{L0}$ as –

$$residual\ filter = base\ filter \times 2^{max(A)-1} \quad (11)$$

So, if the base filter of a level is 32 and $A = 0,1,2,3$, then the residual filter for $C(\cdot)$ will be $32 \times 2^{3-1} = 128$.

**Models.** We develop three variants of U-Net architecture. The properties of each model are tabulated in Table 1. Variants are designed to reduce the no. of model parameters. Architectural details are demonstrated in Supplementary Figure S3-5.

i. Single Recurrent R2U-Net (S-R2U-Net): This model turns on the switch $SW_C$ at $s_{L1}$ and $SW_R$ at $s_{L2}$, thus enabling recurrent at the last set of a level only. The base filters used in this architecture are $\{64, 128, 256, 512, 1024\}$.
ii. Single Recurrent Filter Double R2U-Net (S-R2F2U-Net): Like S-R2U-Net, this model enables recurrent only at $s_{L2}$, while enabling the filter double module. The base filters used in this architecture are $\{32, 64, 128, 256, 512\}$. These reduced base filters significantly reduce the no. of model parameters. Both S-R2U-Net and S-R2F2U-Net disable the attention switch, $SW_A$ in the concatenation path between the decoder and encoder.
iii. Single Recurrent Attention Enabled Filter Double (S-R2F2-Attn-U-Net): It is similar to S-R2F2U-Net except that it enables the attention module in the concatenation path by turning on the $SW_A$ switch.

**Training and inference.** Models are trained and tested on Google Colab Pro. It allows using of up to 15 GB of GPU memory based on availability. The Keras implementation is leveraged from Yingkai Sha's[26] repository. Weight update is done using Adam optimizer[27] with an initial learning rate of 0.001 to reduce losses. Factor 0.1 is set to reduce the learning rate with patience of 5 epochs. Labels are converted to one-hot labels before applying the loss function. Due to the memory constraint, the batch size is set to 2. Models are trained from 10 to 25 epochs. Alom et al.[24] used binary cross-entropy loss to train their R2U-Net model. In this paper, we use a hybrid loss function which is defined as follows –

$$\mathcal{L}_{Total} = \lambda_1 \mathcal{L}_{CE} + \lambda_2 \mathcal{L}_D \quad (12)$$

Where $\mathcal{L}_{CE}$ and $\mathcal{L}_D$ are cross-entropy loss and Sorensen dice loss, respectively. $\lambda_1$ and $\lambda_2$ are two constants and set empirically.

|   | Model | Acc. | Spec. | Pre. | Recall | Dice | Param (M) |
|---|---|---|---|---|---|---|---|
| a. | Attention UNET[22] | 97.06 | 98.60 | 94.28 | 91.22 | 92.55 | 43.94 |
|   | U$^2$ NET[23] | 96.82 | **98.84** | **95.12** | 89.01 | 91.783 | 60.11 |
|   | R2U-NET[24] (up to 4 levels) | 97.27 | 98.66 | 94.61 | 92.02 | 93.11 | 108.61 |
|   | ResUNET-a[25] | 97.10 | 98.50 | 93.95 | 91.77 | 92.66 | 4.71 |
| b. | U-Net[13] | 96.04 | 97.68 | 89.89 | 90.18 | 89.33 | 31.04 |
|   | BiseNet[28] | 95.05 | 95.98 | 85.53 | 92.48 | 87.8 | 12.2 |
|   | DenseASPP[29] | 95.5 | 97.76 | 90.09 | 86.88 | 88.13 | 46.16 |
|   | SegNet[30] | 96.38 | 98.32 | 92.26 | 89.05 | 90.15 | 29.44 |
|   | BASNet[31] | 96.77 | 98.64 | 94.56 | 90.11 | 92.12 | 87.06 |
|   | TSASNet[18] | 96.94 | 97.81 | 94.77 | **93.77** | 92.72 | 78.27 |
|   | MSLPNet[20] | 97.30 | 98.45 | 93.35 | 92.97 | 93.01 | - |
|   | LTPEDN[21] | 94.32 | - | - | - | 92.42 | - |
| c. | S-R2U-Net | 97.22 | 98.52 | 94.06 | 92.33 | 93.01 | 77.17 |
|   | S-R2F2U-Net | **97.31** | 98.55 | 94.27 | 92.61 | **93.26** | 59.12 |
|   | S-R2F2-Attn-U-Net | 97.23 | 98.55 | 94.19 | 92.16 | 93.00 | 59.25 |

**Table 2.** Performance analysis. (a) State-of-the-art models that we explored. (b) some reported state-of-the-art results. (c) our proposed models.

| Category / Model | 1 | 2 | 3 | 4 | 5 | 6 | 7 | 8 | 9 | 10 |
|---|---|---|---|---|---|---|---|---|---|---|
| MSLPNet | **93.98** | **95.16** | 93.38 | **95.96** | 90.39 | 92.12 | 84.29 | **93.93** | 94.89 | 92.5 |
| S-R2U-Net | 93.46 | 95.15 | **95.28** | 95.08 | **94.78** | 93.10 | **85.82** | 91.83 | **95.35** | 94.57 |
| S-R2F2U-Net | 93.56 | 95.07 | **95.28** | 95.10 | **94.78** | 93.30 | **85.82** | 92.18 | 95.29 | **94.70** |
| S-R2F2-Attn-U-Net | 93.28 | 94.92 | 95.05 | 95.03 | 94.39 | **93.36** | 85.80 | 92.31 | 95.27 | 94.34 |

**Table 3.** Dice scores for individual categories.

| Parameter / Model | $\lambda_1$ | $\lambda_2$ | Acc. | Spec. | Pre. | Recall | Dice |
|---|---|---|---|---|---|---|---|
| S-R2U-Net | 1 | 0 | 97.22 | 98.52 | 94.06 | 92.33 | 93.01 |
| S-R2F2U-Net | 1 | 0 | 97.24 | 98.46 | 93.93 | 92.58 | 93.07 |
|   | 1 | 0.5 | 97.31 | 98.55 | 94.27 | 92.61 | 93.26 |
| S-R2F2-attn-U-Net | 1 | 0 | 97.21 | 98.78 | 94.97 | 91.29 | 92.92 |
|   | 1 | 0.5 | 97.31 | 98.50 | 94.00 | 92.77 | 93.22 |

**Table 4.** Analysis of hybrid loss function.

**Evaluation metric.** The dice coefficient is widely used in the medical image segmentation community to evaluate the performance of an image segmentation model[18]. In addition, we use accuracy, specificity, precision, and recall to evaluate our model. Each definition is as follows:

$$Accuracy = \frac{TP + TN}{TP + FP + TN + FN} \quad (13)$$

$$Specificity = \frac{TN}{FP + TN} \quad (14)$$

$$Precision = \frac{TP}{TP + FP} \quad (15)$$

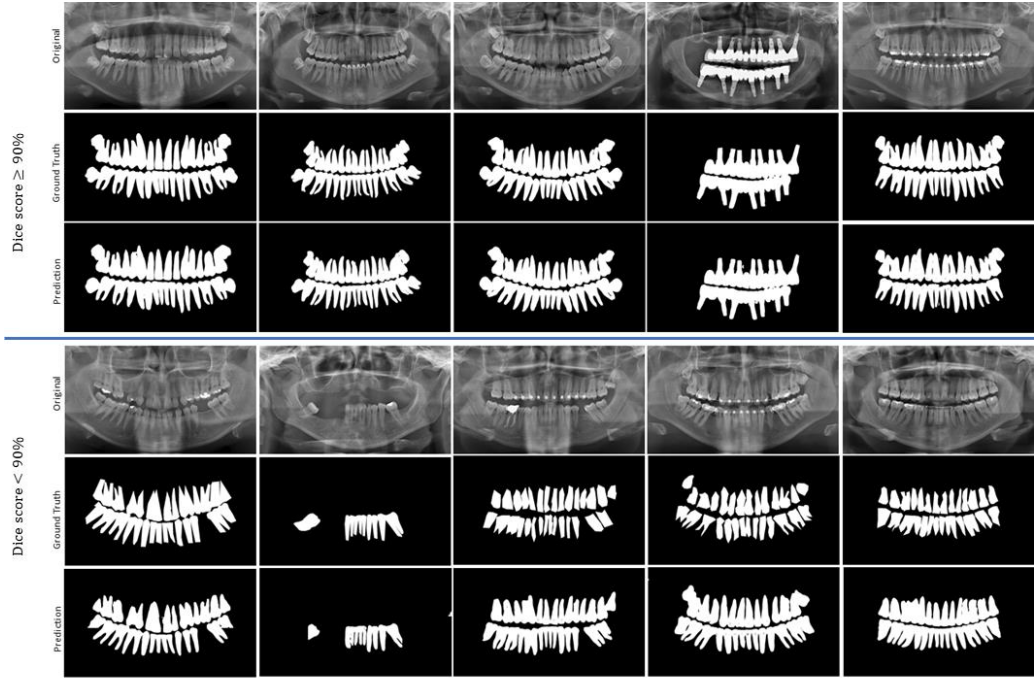

**Figure 2.** Segmentation results for S-R2F2U-Net.

$$Recall = \frac{TP}{TP + FN} \qquad (16)$$

$$Dice = \frac{2|P \cap G|}{|P| + |G|} = \frac{2TP}{2TP + FP + FN} \quad (when\ applied\ to\ boolean\ data) \qquad (17)$$

Where, $TP$: true positive, $TN$: true negative, $FP$: false positive, $FN$: false negative, $P$: predicted region, $G$: ground truth.

## Results

The performance of R2U-Net and Attention U-Net along with the other state-of-the-art methods we explored is tabulated in Table 2(a). As both architectures demonstrate promising performance, we move forward to build our model combining recurrent, residual, and attention with U-Net architecture. The goal is to implement a model which not only obtains a good dice score but is also relatively smaller in size. The entire network architecture and switching mechanism are shown in Figure 1. Table 1 lists some of the configurations that we explore for teeth segmentation. The performance of the proposed models is tabulated in Table 2(c). We notice that the network size can be reduced significantly using the filter doubling technique as described in the previous section while keeping the dice score at a decent level. As indicated in Table 2(b-c), our proposed model achieves the highest dice score compared to the reported state-of-the-art results. In addition, performance analysis for individual categories is listed in Table 3. It is observed that all but category 7 have achieved dice scores of more than 90%. Poor annotation in category 7 is one of the reasons for that. Further analysis is described in the following section. A visual presentation of the segmentation results of the proposed model is shown in Figure 2. The upper half is for Dice scores higher than 90%, whereas the lower half is for dice scores below 90%. Dice scores below 90% mostly occur in category 7. Table 4 demonstrates how the hybrid loss function boosts performance. Figure 3 demonstrates the decoding process of our proposed model. Some complicated cases where the proposed model performed ill are shown in Figure 4(A). A boxplot implementation is shown in Figure 4(B) to summarize the performance of the proposed model.

## Discussion

Our proposed model is built-up on U-Net-based architecture. At first, we explore some state-of-the-art methods to evaluate performance on teeth segmentation from panoramic X-ray images. Promising results, especially from the R2U-Net, are tabulated in Table 2(a). But to achieve this performance, we use an extensively large network architecture for R2U-Net (over 100M parameters) as indicated in Table 2(column–'Param (M)'). It happened for two reasons. First, it uses 18 recurrent modules. Though recurrent modules are good at integrating context information, each recurrent module is very costly in terms of model parameters.

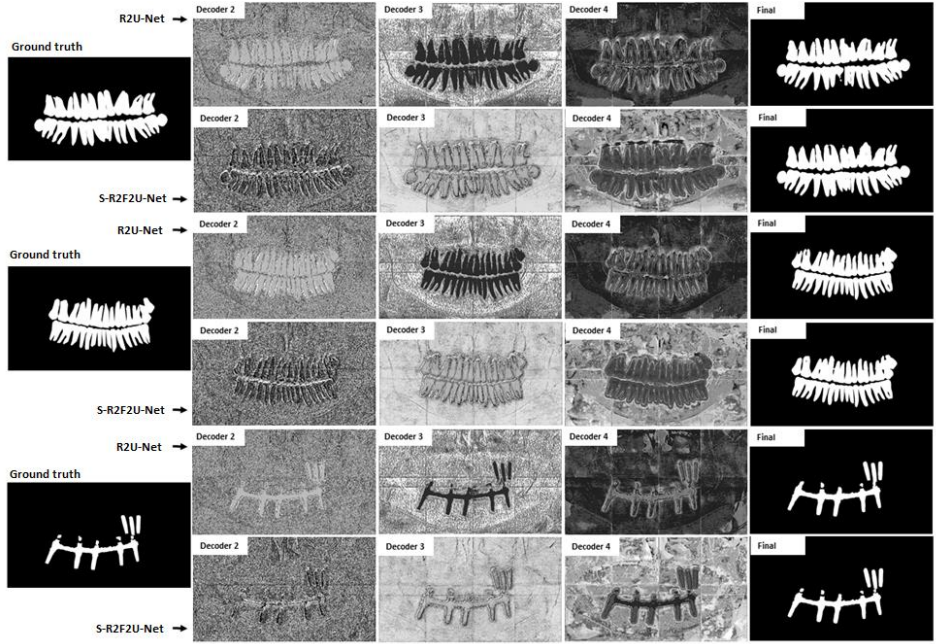

**Figure 3.** Illustration of the decoding process in R2U-Net and S-R2F2U-Net.

Second, base filters used are quite large {64, 128, 256, 512, 1024}. If we consider the encoder path alone, the summation of $f_{L1}$ and $f_{L2}$ end up with a minimum of 3968 feature maps even if we consider Conv-BN-ReLU modules instead of recurrent modules.

As a solution to the first problem, instead of applying recurrent modules to both $s_{L1}$ and $s_{L2}$, we apply it to $s_{L2}$ only, thus reducing 50% of recurrent modules. Our first proposed model, S-R2U-Net, is the product of this. However, as shown in Table 2(c), the dice score is slightly lower than the R2U-Net. To tackle this problem, we introduce S-R2F2U-Net which has two more modifications – filter doubling and hybrid loss function. S-R2U-Net uses the same number of filters in both $s_{L1}$ and $s_{L2}$. Instead, now filters are double at $s_{L2}$ and base filters are lowered to {32, 64, 128, 256, 512}. This way the summation of $f_{L1}$ and $f_{L2}$ in the encoder end up with a minimum of 2976 feature maps. Alom et al.[24] used binary cross-entropy loss in their R2U-Net architecture. In our paper, instead, we used a hybrid loss function that consists of cross-entropy loss and dice loss with two tunable parameters $\lambda_1$ and $\lambda_2$. Experimentally, we have found that $\lambda_1 = 1$ and $\lambda_2 = 0.5$ best work for the model.

As Table 2(a-c) indicates, S-R2F2U-Net achieves a better dice score (93.26%) than R2U-Net (93.11%) and outperforms the state-of-the-art models. The performance analysis of our hybrid loss function is shown in Table 4. Each section's performance improves after the hybrid loss function is introduced. Some of the S-R2F2U-Net outputs are shown in Figure 2. Dice scores less than 90% mostly occurred in category-07. It is the most challenging among all categories to predict due to its poor annotation. Table 2(column–'Param (M)') illustrates that S-R2F2U-Net architecture has 59.12 million model parameters which is less than around 45% as compared to the R2U-Net, and even less than the state-of-the-art TSASNet[18] (78.27M). The fact that TSASNet is a two-stage model, with the first stage extracting contextual information and the second stage performing segmentation, is one reason for this. Unlike TSASNet, our model is a single-stage model that takes advantage of the recurrent module to collect information from nearby units. Performance for individual categories is shown in Table 3. It is clear that, despite its smaller architecture, the S-R2F2U-Net can achieve similar, if not better, results than the R2U-Net. Even our proposed model achieves a higher dice score (85.82%) for the challenging category-07 compared to the state-of-the-art MSLPNet[20] (84.29%). A poor dice score (90.39%) is also reported in MSLPNet for category-5 due to the presence of metal artifacts and metal fillings in the teeth. On the other hand, as shown in Table 3, both R2U-Net and S-R2F2U-Net can tackle much better in such cases with a dice score of 94.78%.

Figure 3 shows the decoding process for both the R2U-Net and S-R2F2U-Net. For better visualization, we resize all the feature maps to the original image size. At decoder-2, models roughly identify the teeth regions, though they are not completely separable from the surroundings. At decoder-3, as the first notable distinction is detected, the effect of context modulation appears. R2U-Net shows a significant intensity difference between the teeth and the surrounding area. S-R2F2U-Net, in contrast, shows no significant differences in pixel intensity. The visible tooth borders, instead, isolate the teeth regions from the rest of the surroundings. At decoder-4, models fine-tune the teeth border regions. A strong intensity difference is seen in S-R2F2U-Net at this stage. In addition, we explore the effect of the attention gate in skip connections. The model S-R2F2-Attn-U-Net uses attention modules in its skip connections. We observe no significant improvement offered by attention gates. If we look at the decoding process, we notice that attention gates focus on various zones in the teeth region, as illustrated in Supplementary Figure S6. As a result, S-R2F2-Attn-U-Net is projected to perform better when segmenting multi-label tiny objects.

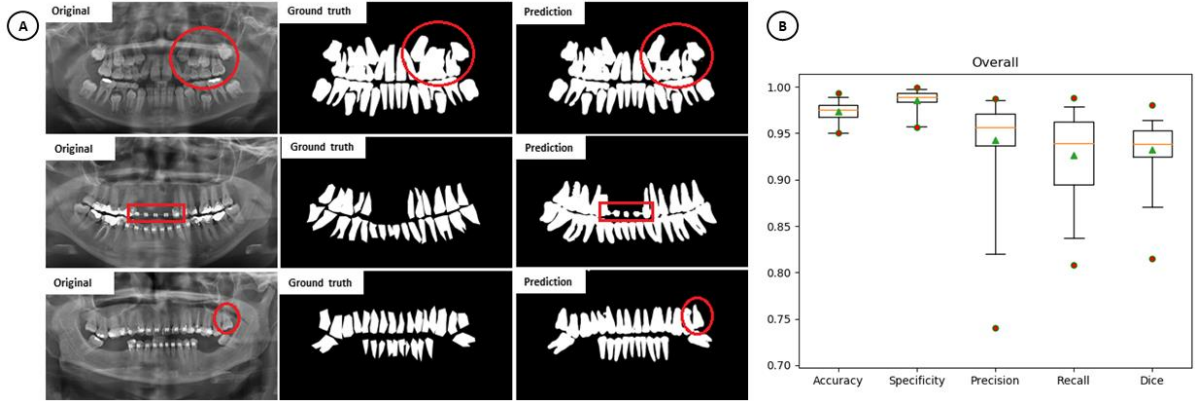

**Figure 4.** More evaluations on S-R2F2U-Net. (A) Examples of some challenging cases. From top to bottom row-wise – effect of supernumerary teeth, foreign boides, and fuzzy tooth roots. (B) Boxplot representation of overall performance.

We find some factors that can influence the model's performance. The presence of supernumerary teeth, foreign bodies, fuzzy dental roots, and poor annotation can all influence segmentation ability in specific situations. supernumerary teeth are not part of the regular tooth series and make the segmentation difficult since there's no clear separation between those teeth, and some of them even overlap. Foreign bodies include dental restorations, broken dental instruments, metal fillings, pins, toothpicks, and other metal objects. However, we haven't noticed a significant impact of foreign bodies on model performance. One exceptional case from category-7 is shown in Figure 4(A, row-2) where the model detects some brackets as the teeth region. It happened because brackets are supposed to attach to teeth, and this is true for all other images. Even the bottom teeth of the same figure have brackets as well. As shown in Figure 4(A, row-3), the tooth root's intensity blends with the intensity of the jaw bone, making separation from the jaw bone difficult. Because such events are uncommon in the dataset, the model hardly encounters them during training. Some categories, especially category-7, are annotated poorly[18,20] and contain sharp edges and cones. Instead, when compared to the original data, our model predicts more smooth edges, which makes it more reasonable. From Figure 4(B), we see the existence of outliers in the boxplot representation indicating some data points exist behind the whiskers. Boxplots for individual categories are shown in Supplementary Figure S7.

## Conclusion

In this paper, we propose three single-stage deep learning models for teeth segmentation from panoramic X-ray images. First, we conduct comprehensive experiments and analyses on 4 state-of-the-art methods (Attention U-Net, R2U-Net, $U^2$-Net, and ResUNet-a) that lead us to propose 3 novel models – Single Recurrent R2U-Net (S-R2U-Net), Single Recurrent Filter Double R2U-Net (S-R2F2U-Net), and Single Recurrent Attention Enabled Filter Double (S-R2F2-Attn-U-Net). Especially, S-R2F2U-Net not only outperforms the state-of-the-art results but also reduces the model parameters by 45% as compared to the original R2U-Net. The model's performance is improved by adopting a hybrid loss function. Future work includes segmenting and labeling each tooth individually.

# S-R2F2U-Net: A single-stage model for teeth segmentation

Mrinal Kanti Dhar[1,*] and Mou Deb[2]

[1]Department of Computer Science, University of Wisconsin-Milwaukee, WI, USA
[2]Department of Electrical Engineering, South Dakota School of Mines and Technology, SD, USA

[1,*]mdhar@uwm.edu and [2]mou.deb@mines.sdsmt.edu

## Supplementary results and discussion

The benchmark dataset used in this paper is summarized in Supplementary Table S1. There are 1500 dental images in the collection, which are grouped into ten categories. Supplementary Figure S1 is a higher-resolution reconstruction of the modules depicted in Figure 1(B-D) in the main manuscript. Supplementary Figure S2 illustrates R2U-Net. It also includes filters used at each level. Supplementary Figures S3-S5 illustrate our 3 proposed models – Single Recurrent R2U-Net (S-R2U-Net), Single Recurrent Filter Double R2U-Net (S-R2F2U-Net), and Single Recurrent Attention Enabled Filter Double (S-R2F2-Attn-U-Net). These are the individual configurations extracted from the switch-based U-Net architecture demonstrated in Figure 1 in the main manuscript. Supplementary Figure S6 depicts the decoding process of (S-R2F2-Attn-U-Net). The decoding process of R2U-Net and S-R2F2U-Net is shown in the main manuscript. Category-wise performance obtained by S-R2F2U-Net is illustrated using boxplots in Supplementary Figure S7. In the main manuscript, only the overall performance is depicted in Figure 4(B).

| Category / Attribute | 1 | 2 | 3 | 4 | 5 | 6 | 7 | 8 | 9 | 10 |
|---|---|---|---|---|---|---|---|---|---|---|
| Missing teeth | | | | | ✓ | | ✓ | ✓ | ✓ | ✓ |
| Restoration | ✓ | ✓ | | | | | ✓ | ✓ | | |
| Dental appliance | ✓ | | ✓ | | | | ✓ | | ✓ | |
| Dental implant | | | | | ✓ | | | | | |
| Images | 73 | 220 | 45 | 140 | 120 | 170 | 115 | 457 | 45 | 115 |
| Average teeth | 32 | 32 | 32 | 32 | 18 | 37 | 27 | 29 | 28 | 28 |

**Supplementary Table 1.** Summary of the benchmark dataset.

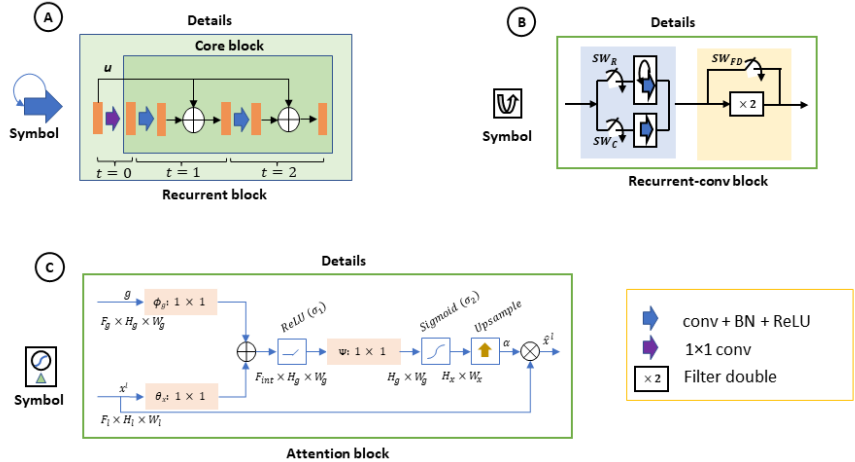

**Supplementary Figure 1.** (A) A recurrent module is illustrated for $t = 2$. The recurrent inputs evolve over iterations. The feedforward unit $u$ remains unchanged during the entire process. It is added to the recurrent inputs in every time step except for $t = 0$. (B) The recurrent-conv block is illustrated. Feature maps can be generated either by applying Conv-BN-ReLU (CBR) or by recurrent. Switches $SW_R$ and $SW_C$ are used to activate recurrent and CBR modules, respectively. Only one of them can be turned on at a time. In addition, the switch $SW_{FD}$ is embedded to enable filter doubling. It doubles the input filters. (C) The attention module is illustrated. It generates additive attention by applying some linear and non-linear operations. The attention output is the elementwise multiplication of the input and the attention coefficient, α.

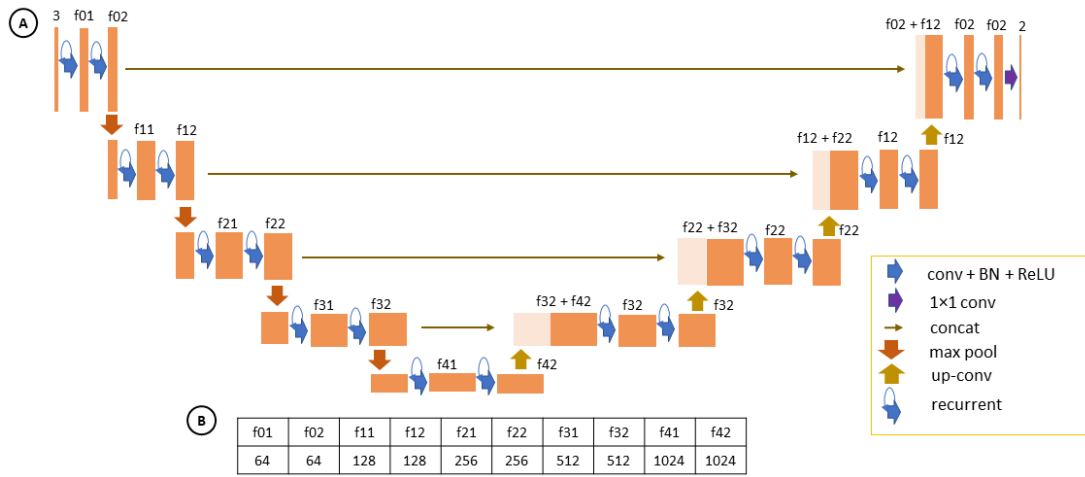

**Supplementary Figure 2.** Illustration of R2U-Net. (A) Network architecture. (B) Chart indicating no. of filters used in each level. Base filters are {64, 128, 256, 512, 1024}. Two recurrent modules are used in each encoder or decoder level. Filter doubling is not used. So, $f_{L2} = f_{L1}$ at the $L^{th}$ level.

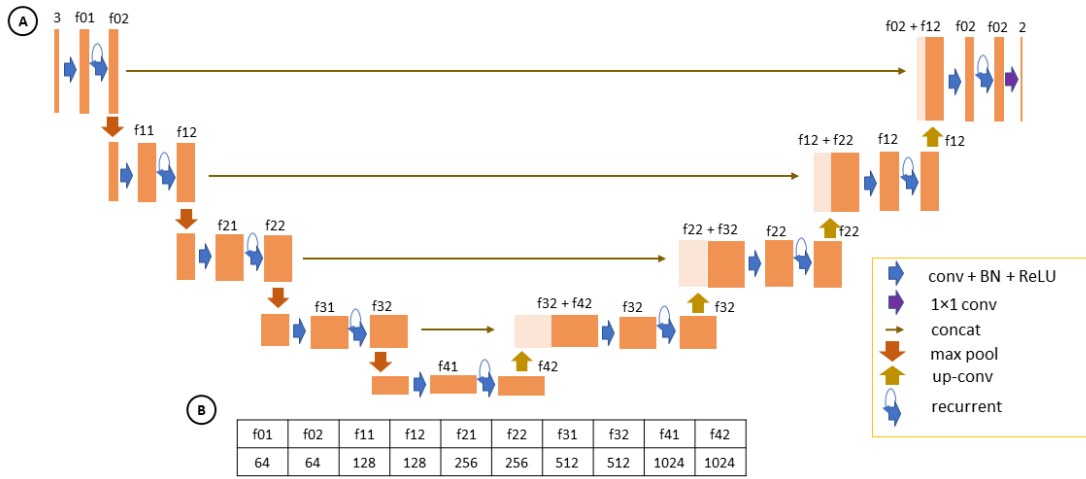

**Supplementary Figure 3.** Illustration of Single Recurrent R2U-Net (S-R2U-Net). (A) Network architecture. (B) Chart indicating no. of filters used in each level. Base filters are {64, 128, 256, 512, 1024}. Instead of two, now only one recurrent module is used in each encoder or decoder level. Filter doubling is not used. So, $f_{L2} = f_{L1}$ at the $L^{th}$ level.

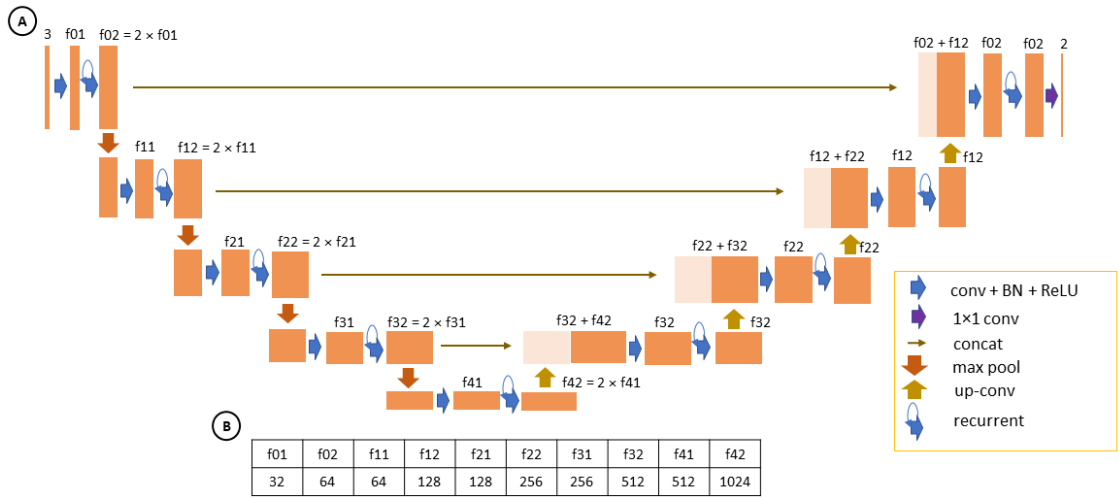

**Supplementary Figure 4.** Illustration of Single Recurrent Filter Double R2U-Net {S-R2F2U-Net}. (A) Network architecture. (B) Chart indicating no. of filters used in each level. Base filters are {32, 64, 128, 256, 512}. Instead of two, only one recurrent module is used at each encoder or decoder level. Filter doubling is enabled now. So, $f_{L2} = 2 \times f_{L1}$ at the $L^{th}$ level.

**Supplementary Figure 5.** Illustration of Single Recurrent Attention Enabled Filter Double (S-R2F2-Attn-U-Net). (A) Network architecture. (B) Chart indicating no. of filters used in each level. Base filters are {32, 64, 128, 256, 512}. Instead of two, only one recurrent module is used at each encoder or decoder level. Filter doubling is enabled now. So, $f_{L2} = 2 \times f_{L1}$ at the $L^{th}$ level. Details of the attention gate are demonstrated in the main manuscript.

**Supplementary Figure 6.** Illustration of the decoding process in Single Recurrent Attention Enabled Filter Double (S-R2F2-Attn-U-Net). As indicated by decoders 3 and 4, it is focusing on the inner part of the tooth region.

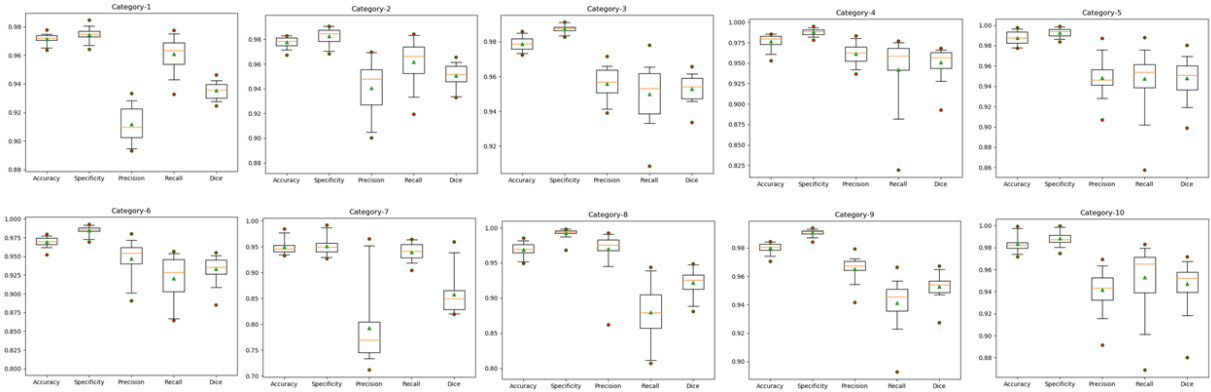

**Supplementary Figure 7.** Boxplot representation of the results obtained by S-R2F2U-Net. The whiskers endpoints are selected as the (1st quartile - 1.5 × interquartile range (IQR)) and (3rd quartile + 1.5 × IQR). The median is represented by an orange line, whereas the mean is represented by a green triangle. The x-axis label shows the evaluation metrics – accuracy, specificity, precision, recall, and dice score. The points that are outside the interval specified by the whiskers are referred to as outliers. Outermost outliers are demonstrated only.